\documentclass[conference]{IEEEtran}
\IEEEoverridecommandlockouts

\usepackage{tikz}
\usepackage{cite}
\usepackage{amsmath,amssymb,amsfonts}
\usepackage{algorithmic}
\usepackage{graphicx}
\usepackage{textcomp}
\usepackage{xcolor}
\usepackage{subcaption}
\usepackage{url}
\def\BibTeX{{\rm B\kern-.05em{\sc i\kern-.025em b}\kern-.08em
    T\kern-.1667em\lower.7ex\hbox{E}\kern-.125emX}}
\begin{document}

\title{Virtual Chromoendscopy with\\Tunable Visibility Enhancement\\
\thanks{This study was supported by AMED under Grant Numbers JP24hma922022 and JP25hma322042.  It was also supported by JSPS KAKENHI under Grant Number JP24K15772.}
}

\author{Yuhi Kanno$^{1}$, Yusuke Monno$^{1}$, Sho Suzuki$^{2}$, Tomohiro Tada$^{3}$, and  Masatoshi Okutomi$^{1}$
\thanks{$^{1}$Y. Kanno, Y. Monno, and M. Okutomi are with the Department of Systems and Control Engineering, School of Engineering, Institute of Science Tokyo, Meguro-ku, Tokyo 152-8550, Japan (email: {ykanno@ok.sc.e.titech.ac.jp}).}
\thanks{$^{2}$S. Suzuki is with the Department of Gastroenterology, International University of Health and Welfare Ichikawa Hospital, Ichikawa-shi, Chiba 272-0827, Japan.}
\thanks{$^{3}$T. Tomohiro is with AI Medical Service Inc., Toshima-ku, Tokyo 170-0013, Japan.}
}

\maketitle

\begin{abstract}
Chromoendoscopy (CE) is a common clinical practice that sprays indigo carmine blue dye onto the gastric surface to improve the visibility of gastric lesions, such as an early cancer. While CE is effective in detecting the lesions, preparing and spraying the dye needs additional cost and time, which is undesirable both for patients and medical practitioners. To overcome this issue, virtual chromoendoscopy (V-CE) was recently proposed, which applies a learned image translation model to virtually generate a CE image from a standard endoscopy (SE) image. In this paper, we propose virtual \textit{enhanced chromoendoscopy} (\mbox{V-ECE}) that combines V-CE with image enhancement techniques to further improve the visibility of gastric lesions. Because a desired enhancement level depends on the inspected lesion and the practitioner's preference, we introduce a novel image translation model that can generate V-ECE images using an enhancement level tunable by a user. Experimental results demonstrate that our proposed model can plausibly generate V-ECE images with various enhancement levels using a unified model.

\end{abstract}

\begin{IEEEkeywords}
Endoscope, chromoendoscopy, image translation, image enhancement, generative adversarial network
\end{IEEEkeywords}

\section{Introduction}

\begin{figure}[t!]
    \centering
    \includegraphics[width=1\linewidth]{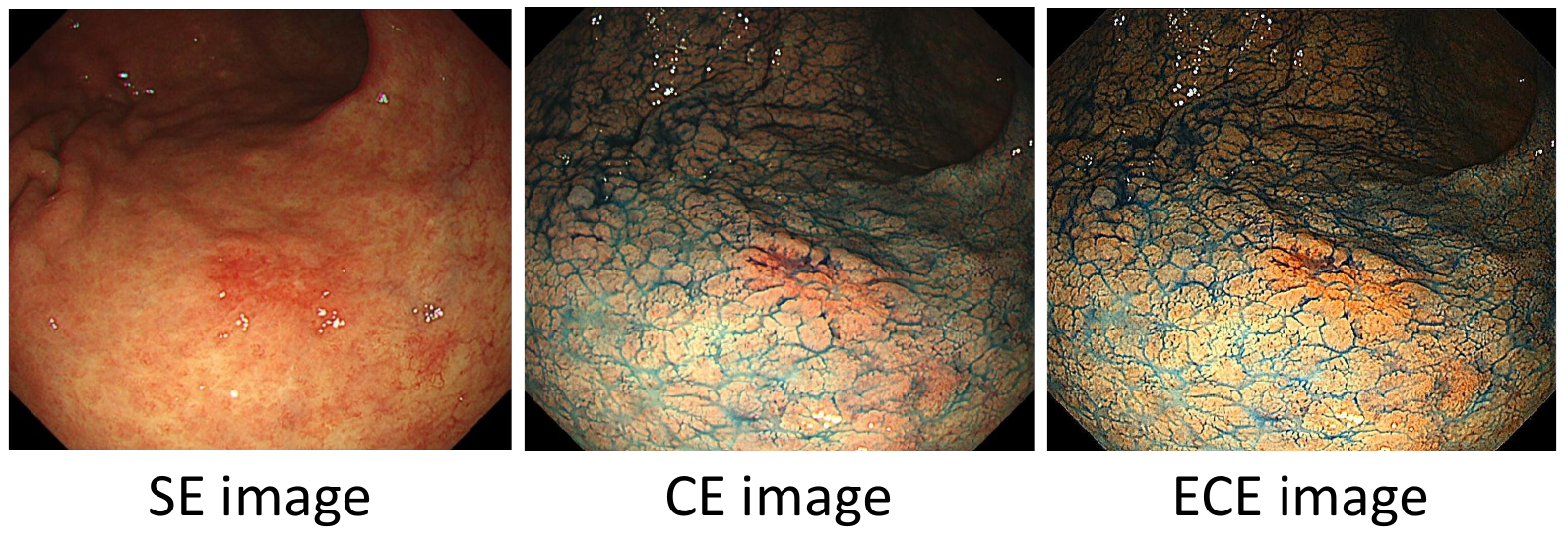}
    \caption{The comparisons of a standard endoscopy (SE) image, a chromoendoscopy (CE) image, and an enhanced chromoendoscopy (ECE) image.}
    \label{fig:example}
\end{figure}

Chromoendoscopy (CE) is a well-adopted clinical practice to improve the visibility of gastric lesions, such as early cancer, by spraying indigo carmine blue dye from the endoscope tip onto the gastric surface. As shown in Fig.~\ref{fig:example}, the CE image shows better lesion visibility (around the center of the image) than the standard endoscopy (SE) image, because the blue dye enhances surface structure and color contrast.  While CE is effective in assisting the detection of early lesions~\cite{zhao2016meta}, preparing and spraying the blue dye during endoscopy requires additional cost and time, which is undesirable not only for patients but also for medical practitioners. This prevents practitioners from performing CE proactively and frequently.

\begin{figure}[t!]
    \centering
    \includegraphics[width=1\linewidth]{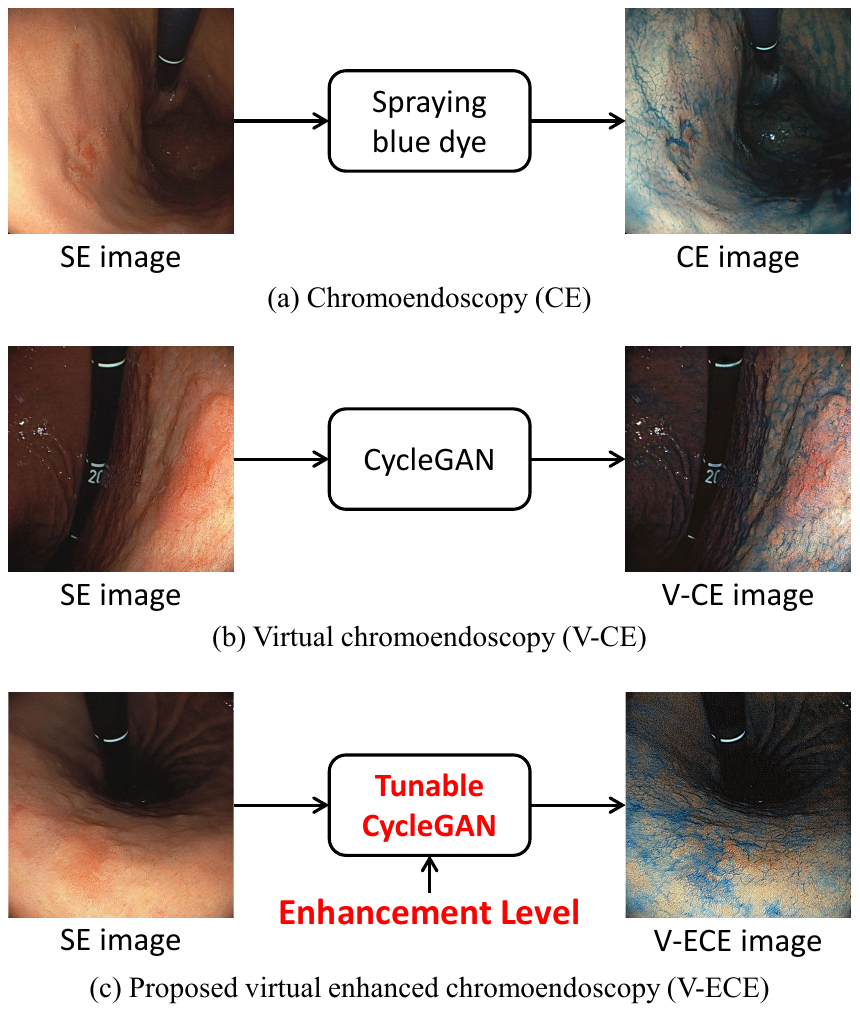}
    \caption{The technical differences among CE, V-CE~\cite{suzuki2024diagnostic}, and our proposed V-ECE. Our proposed method generates a V-ECE image, where the enhancement level is tunable by a user.\label{fig:concept}}
    \label{fig:Comparison}
\end{figure}

\begin{figure*}
    \centering
    \begin{minipage}{0.66\textwidth}
        \centering
        \includegraphics[width=1.0\linewidth]{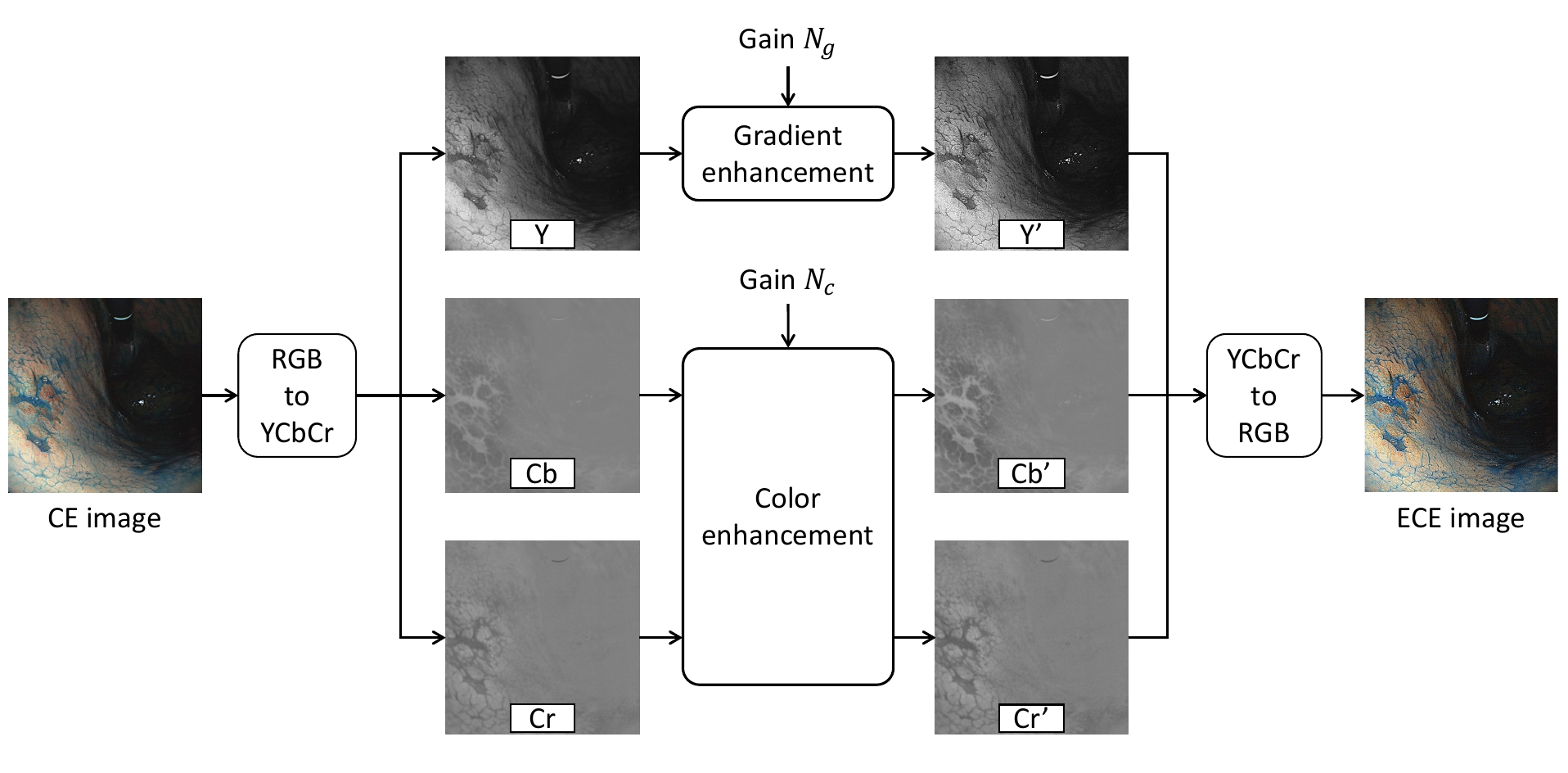}\\
        \vspace{-2mm}
        \subcaption{Overall flow of our visibility enhancement}
    \end{minipage}
    \begin{minipage}{0pt}
      \begin{tikzpicture}
        \hspace{2mm} \draw[dashed] (0,-1cm) -- (0,5cm);
      \end{tikzpicture}
    \end{minipage}
    \begin{minipage}{0.32\textwidth}
        \centering
        \vspace{3mm}
        \includegraphics[width=0.82\linewidth]{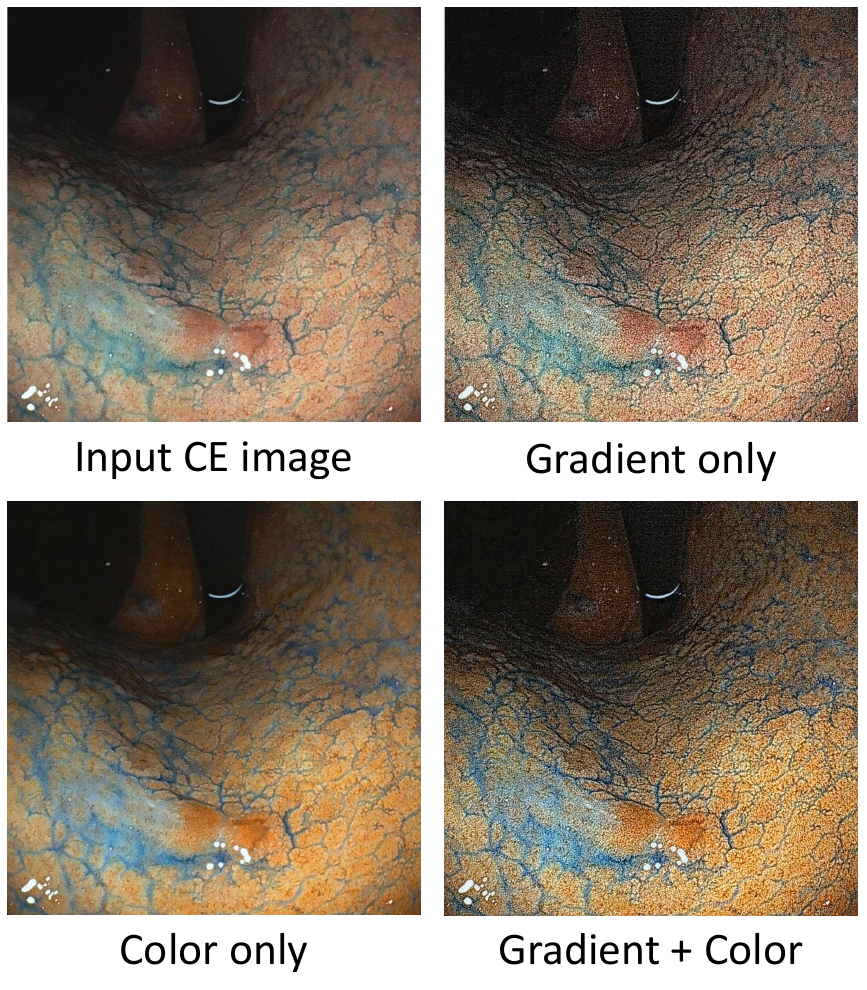}\\ \vspace{-1mm}
        \subcaption{Enhancement results}
    \end{minipage}
    \caption{(a) The overall flow of our visibility enhancement method. We apply gradient enhancement to the Y component and color enhancement to the Cb and Cr components. (b) The examples of enhancement results. The gradient gain $N_g$ and the color gain $N_c$ are set to 5.0 and 1.2, respectively.}
    \label{fig:image_enhancement}
\end{figure*}
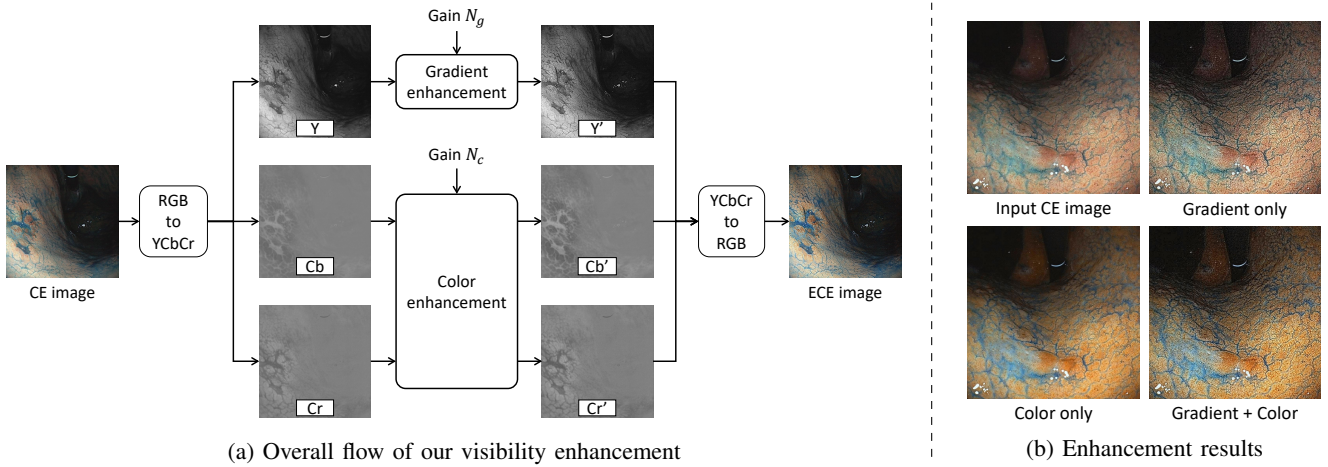

To overcome the above-mentioned issue, our previous studies~\cite{widya2021stomach,suzuki2024diagnostic,takasu2025assessment,suzuki2026generative} and the other study~\cite{fukuda2019generating} proposed a virtual chromoendoscopy (V-CE) technique. As shown in Figs.~\ref{fig:concept}(a) and~\ref{fig:concept}(b), V-CE eliminates the necessity of physically spraying the blue dye by applying a learned image translation model, typically a cycle-consistent generative adversarial network~(CycleGAN~\cite{zhu2017unpaired}), to virtually generate a CE image from an SE image. In our clinical studies, we reported that V-CE demonstrates better visibility for gastric neoplasms and early gastric cancers compared with SE~\cite{suzuki2024diagnostic,takasu2025assessment,suzuki2026generative}. We also confirmed the feasibility of V-CE by developing a real-time V-CE system working with a real endoscope setup~\cite{takasu2025assessment,suzuki2026generative}.

In this paper, we aim to further improve the lesion visibility by combining V-CE with image enhancement techniques. We first present our image enhancement method consisting of gradient enhancement and color enhancement. Figure~\ref{fig:example} shows an example of the comparison between a CE image and an enhanced chromoendoscopy (ECE) image by our method. We can confirm that the ECE image demonstrates better visibility for the lesion existing around the center of the image. Motivated by this, we introduce virtual enhanced chromoendoscopy (V-ECE), which learns a CycleGAN model that translates an SE image to the ECE image, as illustrated in Fig.~\ref{fig:concept}(c)\footnote{Throughout the paper, we use SE, CE, and ECE to present real images, and use V-CE and V-ECE to present virtual images.}. In real clinical practice, a desired image enhancement level depends on the inspected lesion and the medical practitioner's preference. Thus, we propose a novel image translation model named tunable CycleGAN that can generate a V-ECE image using an enhancement level tunable by a user. By this translation model, the practitioner can adjust the image enhancement level according to real endoscopy inspection. We experimentally validate that our tunable CycleGAN model can generate plausible V-ECE images with provided enhancement levels by a single unified model and provide consistent results compared with the dedicated models learned for each enhancement level.

\section{Virtual Enhanced Chromoendoscopy (V-ECE)}

In this section, we introduce V-ECE, which learns a translation model to convert an SE image to a V-ECE image. V-ECE consists of two steps. (i)~Visibility enhancement: This step applies image enhancement techniques to CE images to generate the corresponding ECE images to improve the visibility. (ii)~CycleGAN training: This step learns a CycleGAN model using the unpaired data of SE images~(domain $X$) and the generated ECE images~(domain $Y$). Each step is detailed in Sec.~\ref{ssec:ve} and Sec.~\ref{ssec:cyclegan}. Then, we further propose a tunable CycleGAN model that makes the enhancement level of V-ECE tunable by a user in the application phase, which is detailed in Sec.~\ref{ssec:tunablegan}.

\subsection{Visibility Enhancement} \label{ssec:ve}

\subsubsection{Overview\label{ssec:overview}} We apply two types of image enhancement: gradient and color enhancements. This is based on the following observations: (i)~The human visual perception is more sensitive to pixel value differences, i.e., image gradients, than pixel values themselves. Thus, the visibility of subtle surface structures visualized by the blue dye is expected to be improved by applying the gradient enhancement. (ii)~Gastric lesions typically become reddish in color, and their borders are made easier to detect by the blue dye. Thus, the visibility of the lesion is expected to be improved by applying the color contrast enhancement between reddish and bluish colors.

Figure~\ref{fig:image_enhancement}(a) shows the overview of our visibility enhancement method. It first converts a CE image from the RGB color space to the YCbCr color space~\cite{poynton1996technical} as

\begin{align}
    \begin{bmatrix}
        Y\\Cb\\Cr
    \end{bmatrix}
    &=\begin{bmatrix}
        0.257&0.504&0.098\\
        -0.148&-0.291&0.439\\
        0.439&-0.368&-0.071
    \end{bmatrix}\begin{bmatrix}
        R\\G\\B
    \end{bmatrix}+\begin{bmatrix}
        16\\128\\128
    \end{bmatrix},
    \label{eq:rgb2ycbcr}
\end{align}
where Y represents luminance, and Cb and Cr represent chrominance. We apply the gradient enhancement to the Y component and the color enhancement to the CbCr component, because image gradients and colors are encoded in Y and CbCr, respectively. After the enhancements, the image is returned to the RGB space by the reverse process of Eq.~(\ref{eq:rgb2ycbcr}).

\subsubsection{Gradient enhancement} We apply the image reconstruction method of~\cite{Shibata_2016_CVPR}. For simplicity, we consider uniformly enhancing the gradients for every pixel and provide the target gradient to the method~\cite{Shibata_2016_CVPR} as
\begin{equation}
    q_d(i,j)=N_g \cdot \partial_d u(i,j),
    \label{eq:gradient_operation}
\end{equation}
where $u(i,j)$ is the pixel value of the pixel position $(i,j)$ and $\partial_d$ represents a partial derivative along the direction $d \in \{h,v\}$ to derive the image gradients in the horizontal~($h$) and the vertical~($v$) directions. $q_d(i,j)$ is the target gradient, which is the multiplication of the original gradient by the gain factor~$N_g$. With this target gradient, the method~\cite{Shibata_2016_CVPR} reconstructs the gradient-enhanced image through an optimization process.

\subsubsection{Color enhancement} Color enhancement is designed to enhance red-blue color contrast. By rearranging Eq.~(\ref{eq:rgb2ycbcr}), we obtain
\begin{align}
    \begin{bmatrix}
        Y-16\\Cb-128\\Cr-128
    \end{bmatrix}
    &=\begin{bmatrix}
        0.257&0.504&0.098\\
        -0.148&-0.291&0.439\\
        0.439&-0.368&-0.071
    \end{bmatrix}\begin{bmatrix}
        R\\G\\B
    \end{bmatrix}.
    \label{eq:rgb2ycbcrw/offset}
\end{align}
According to this equation, Cb and Cr with the offset, i.e., $Cb-128$ and $Cr-128$, contain red-blue differences, i.e., $-0.148R+0.439B$ and $0.439R-0.071B$, respectively. Thus, increasing these differences enhances the red-blue color contrast. However, independently performing it requires two gain parameters, which prevents intuitive parameter control. Thus, we introduce a single gain model.

Firstly, the Cr component with the offset (the left side of Eq.~(\ref{eq:rgb2ycbcrw/offset})) is simply multiplied by the color gain $N_c$ to increase the red-blue difference as
\begin{align}
    Cr'-128&= N_c \cdot (Cr-128) \label{eq:cr'w/offset}.
\end{align}
From this equation, we calculate the enhanced Cr as
\begin{align}
    Cr'= N_c \cdot (Cr-128)+128. \label{eq:cr'}
\end{align}
Then, we design the enhanced Cb so that the G value among the RGB values remains constant as an anchor before and after the color enhancement as
\begin{align}
    Cb'= \alpha(N_c-1)(Cr-128) + Cb,
    \label{eq:Cb_modify}    
\end{align}
where $\alpha \approx 2.08$. By doing this, the color enhancement can be performed using the single gain parameter in an interpretable manner. The derivation of the fixed constant~$\alpha$ is described in the Appendix.

Figure~\ref{fig:image_enhancement}(b) shows examples of our enhancement results. We can confirm that our enhancement method combining the gradient and the color enhancements effectively improves the visibility of the gastric surface.

\subsection{CycleGAN Training} \label{ssec:cyclegan}

Because the endoscope or the gastric surface inevitably move during spraying the blue dye, the pixel-aligned paired data cannot be obtained for SE and CE/ECE images. Thus, we adopt CycleGAN~\cite{zhu2017unpaired}, which is a representative image translation model learnable with the unpaired sets of images in two style domains. Two approaches can be considered for V-ECE image generation using CycleGAN: (i)~Sequential: This approach first applies a CycleGAN model learned to translate an input SE image to a V-CE image and then applies the visibility enhancement method to generate the enhanced result. (ii)~Direct: This approach learns a direct mapping from the SE image to the ECE image by a CycleGAN model. In this study, we adopt the direct approach because of two reasons: (i)~It requires only a one-step process and is simpler and faster than the two-step process of the sequential approach. (ii)~It can be implemented in a real clinical environment by changing the CycleGAN model parameters of our developed real-time V-CE system~\cite{takasu2025assessment,suzuki2026generative} 

\begin{figure}[t!]
    \centering
    \includegraphics[width=1\linewidth]{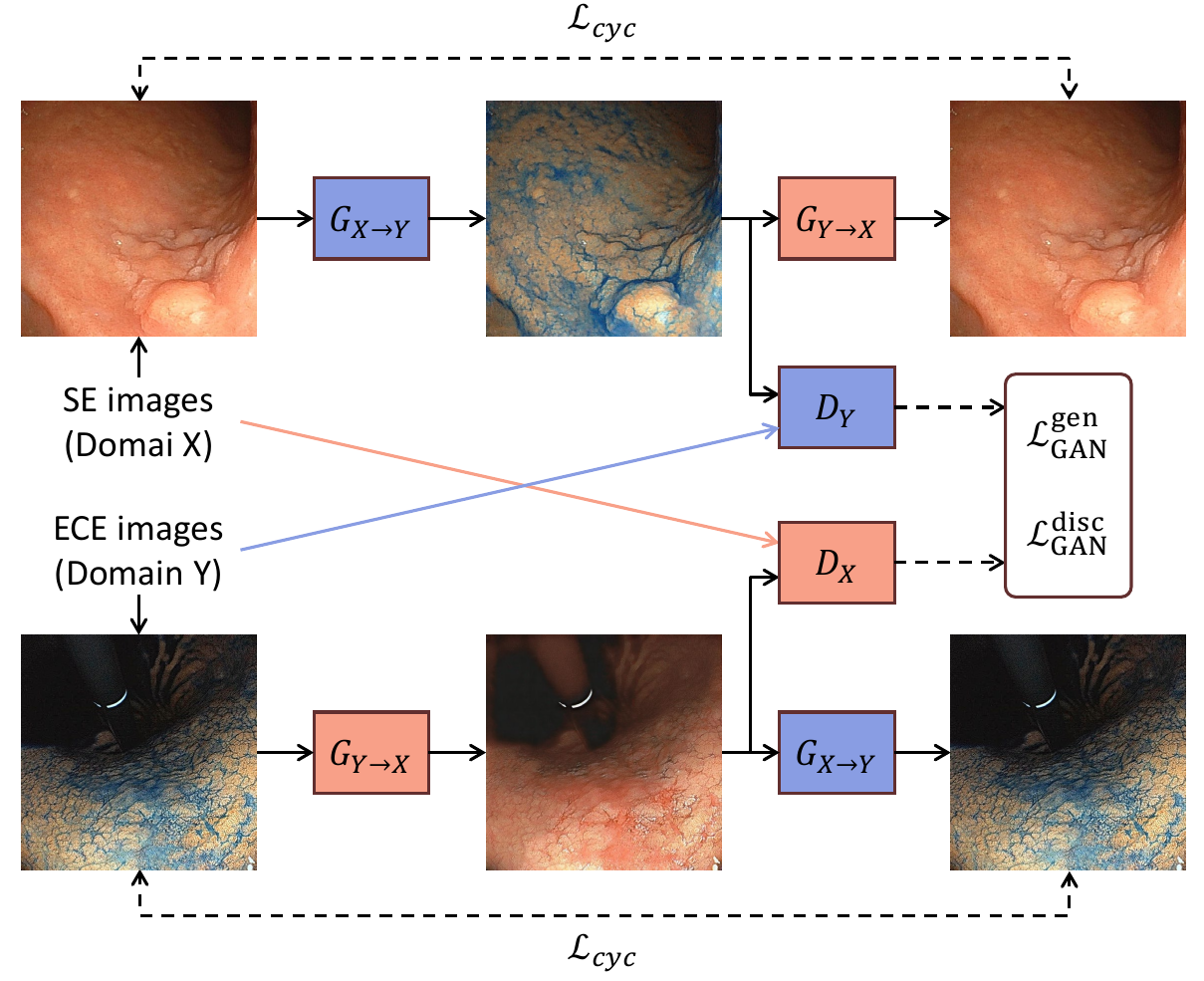}
    \caption{The applied model architecture of CycleGAN.}
    \label{fig:VEIC_method}
\end{figure}

\begin{figure*}[t!]
    \centering
    \includegraphics[width=0.7\linewidth]{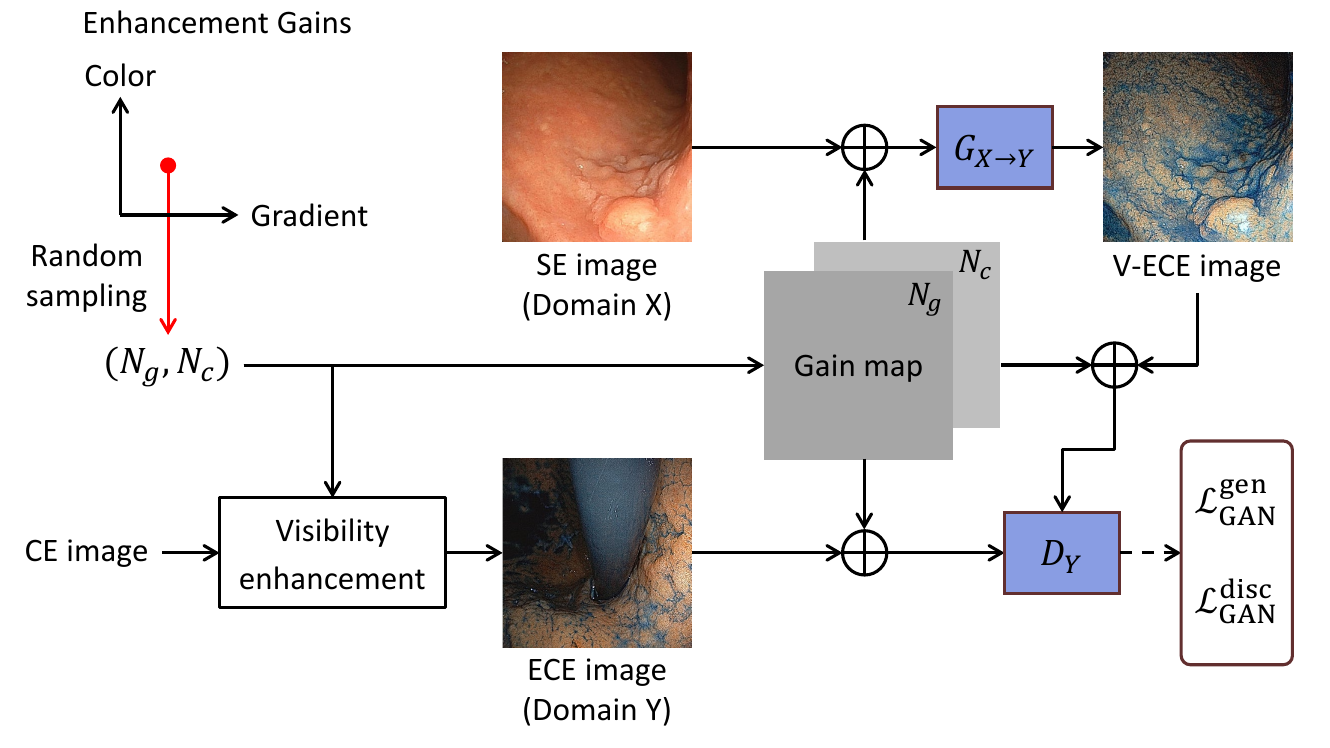}
    \caption{The architecture of our proposed tunable CycleGAN. The gradient and the color gains are randomly sampled from certain ranges and concatenated as the inputs to the generator and the discriminator. The generator tries to generate V-ECE images with the corresponding gains, while the discriminator tries to discriminate them from real ECE images enhanced using the same gains. By this architecture, the discriminator is to identify not only whether the ECE image is real or not, but also whether the enhancement level is correct or not, directing the generator to generate the V-ECE images with the correct enhancement level.}
    \label{fig:T-VEIC_method}
\end{figure*}

Figure \ref{fig:VEIC_method} shows the applied model architecture of CycleGAN. We define the domain $X$ as SE images and the domain $Y$ as ECE images, which are generated as explained in Sec.~\ref{ssec:ve}. The generator $G_{X \rightarrow Y}$ translates the domain from $X$ to $Y$ and vice versa for the generator $G_{Y \rightarrow X}$. The discriminators $D_X$ and $D_Y$ discriminate real samples and generated samples in each domain.

The loss functions for the generators and the discriminators are described as
\begin{align}
    \min_{G_{X \rightarrow Y}, G_{Y \rightarrow X}} &\mathcal{L}^\text{gen}_{\text{GAN}}+\lambda_\text{cyc}\mathcal{L}_{\text{cyc}}+\lambda_\text{identity}\mathcal{L}_{\text{identity}}, \label{eq:lossG} \\
    \min_{D_X, D_Y} \  &\mathcal{L}^\text{disc}_{\text{GAN}}, \label{eq:lossD} 
\end{align}
where we follow the original CycleGAN implementation by the authors~\cite{zhu2017unpaired} with least squares generative adversarial networks~\cite{mao2017least}.

The adversarial loss $\mathcal{L_\text{GAN}}$ is designed so that the generator tries to generate images that fool the discriminator, while the discriminator tries to discriminate them from real images. The loss functions for the pair of the generator $G_{X \rightarrow Y}$ and the discriminator $D_Y$ are as follows.
\begin{equation}
\begin{aligned}
\mathcal{L}^\text{gen}_{\text{GAN}}
    &= \mathbb{E}_{x\sim p_{\text{data}}(x)}[(D_Y(G_{X \rightarrow Y}(x))-1)^2],\\
\mathcal{L}^\text{disc}_{\text{GAN}}
    &= \mathbb{E}_{x\sim p_{\text{data}}(x)}[D_Y(G_{X \rightarrow Y}(x))^2] \\
    &\quad + \mathbb{E}_{y\sim p_{\text{data}}(y)}[(D_Y(y)-1)^2],
\end{aligned}
\label{eq:GANloss}
\end{equation}
where $x\in X$ and $y\in Y$ represent a sample from each domain, and $p_{data}(x)$ and $ p_{data}(y)$ are probability distributions of them, respectively. By jointly learning the generator and the discriminator using these adversarial losses, the generator is encouraged to generate a virtual image close to a real image. The loss functions for the pair of $G_{Y \rightarrow X}$ and the discriminator $D_X$ are defined in the same manner.

The cycle-consistency loss $\mathcal{L}_{\text{cyc}}$ is designed so that the translation from $X$ to $Y$ and the reverse translation from $Y$ to $X$ should be cyclic. The loss function is as follows.
\begin{equation}
\begin{aligned}
\mathcal{L}_{\text{cyc}}
    &= \mathbb{E}_{x\sim p_{\text{data}}(x)}[\|G_{Y \rightarrow X}(G_{X \rightarrow Y}(x))-x\|_1] \\
    &\quad + \mathbb{E}_{y\sim p_{\text{data}}(y)}[\|G_{X \rightarrow Y}(G_{Y \rightarrow X}(y))-y\|_1].
\end{aligned}
\label{eq:cycloss}
\end{equation}
Through this loss, CycleGAN achieves unsupervised learning and eliminates the need for paired data between the domains $X$ and $Y$.

The identity loss $\mathcal{L_\text{identity}}$ is introduced to prevent excessive or unnecessary transformations as
\begin{equation}
\begin{aligned}
\mathcal{L}_{\text{identity}}
    &= \mathbb{E}_{x\sim p_{\text{data}}(x)}[\|G_{Y \rightarrow X}(x)-x\|_1] \\
    &\quad +\mathbb{E}_{y\sim p_{\text{data}}(y)}[\|G_{X \rightarrow Y}(y)-y\|_1],
\end{aligned}
\end{equation}
which regularizes the network training.

\subsection{Tunable CycleGAN Model} \label{ssec:tunablegan}

We here propose a tunable CycleGAN model that makes the enhancement level, i.e., the gradient and color gains, tunable by a user in the application phase. Figure~\ref{fig:T-VEIC_method} shows the model architecture, where we apply simple yet effective modifications to the original CycleGAN. To prepare the ECE image in the domain $Y$, we randomly select the gradient and the color gains $(N_g, N_c)$ from practical ranges for each gain. The selected gain pair is used to enhance a real CE image, which is sampled from the training dataset, by the visibility enhancement method in Sec.~\ref{ssec:ve}. Inspired by the study of~\cite{zhang2018ffdnet}, which provides a noise level map as the network input to achieve flexible denoising capability, we construct two-channel gain maps (constant-value images of the same size as the RGB image) and concatenate them to the original three-channel RGB image to input to the generator $G_{X \rightarrow Y}$ and the discriminator $D_Y$. With this architecture, the generator tries to generate V-ECE images with the corresponding gains, while the discriminator tries to discriminate them from real ECE images enhanced using the same gains. Here, the role of the discriminator is to identify not only whether the ECE image is real or not, but also whether the enhancement level is correct or not, directing the generator to generate the V-ECE images with the correct enhancement level. For the reverse translation, the gain maps are fed as the inputs only for the generator $G_{Y \rightarrow X}$, because the role of the discriminator $D_{X}$ is to identify whether the SE image is real or not, which is irrelevant to the enhancement level. The loss functions for the generators and the discriminators are the same as Eqs.~(\ref{eq:lossG}) and~(\ref{eq:lossD}). In the application phase, we simply input the gains (gain maps) provided by the user to the generator $G_{X \rightarrow Y}$ to generate the V-ECE images with the corresponding enhancement level.

\begin{figure*}[t!]
  \centering
  \includegraphics[width=1\textwidth]{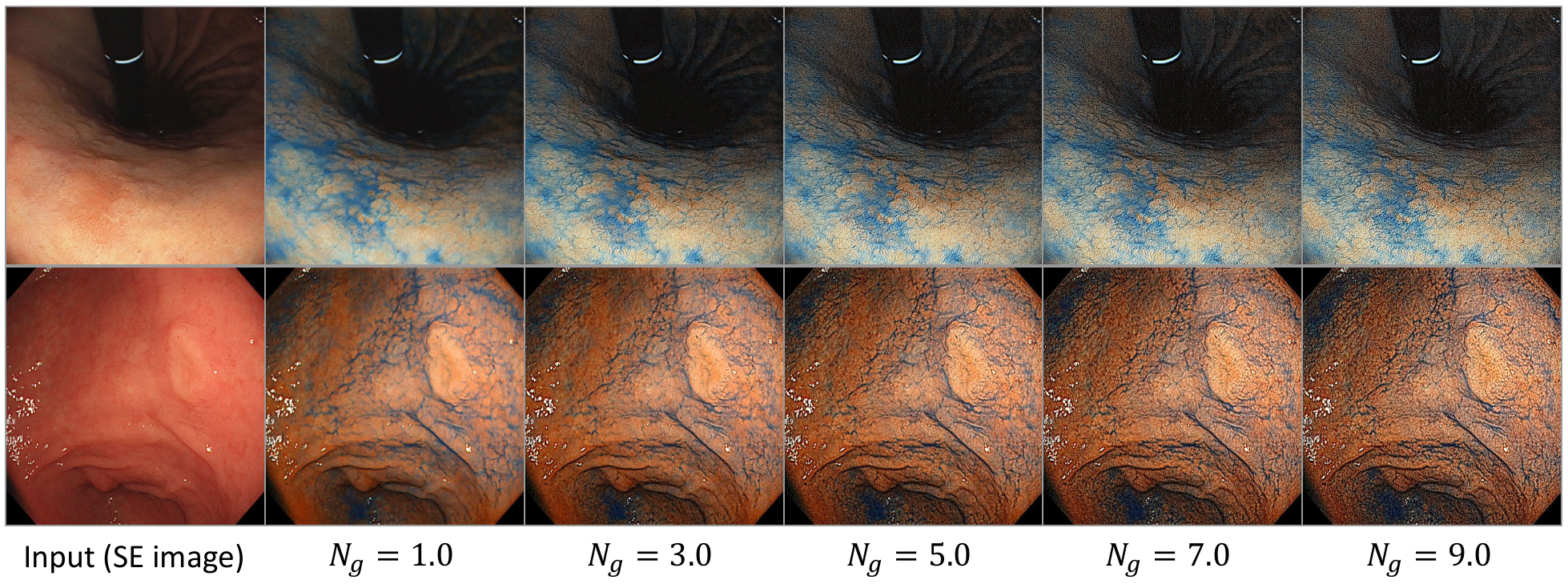}
  \caption{The visual results of the V-ECE image generation by our tunable CycleGAN model. The top row is the results for the Fujifilm dataset, and the bottom row is the results for the Olympus dataset. We can confirm that stronger gradient enhancement is achieved by inputting higher gradient gains.}
  \label{fig:T-VEIC_visual_results}
\end{figure*}

\section{Experimental Results}

\subsection{Ethics}
This study was conducted in accordance with the Declaration of Helsinki. The Institutional Review Board at International University of Health and Welfare approved the study protocol and the data collection on June 25, 2024 (Approval No. 24-Ic-002). Informed consent was obtained from patients before collecting endoscopic images. This study was also approved by the Research Ethics Committee of Institute of Science Tokyo, on July 25, 2024 (Approval No. 2024150) to process the collected data.

\subsection{Datasets}

We constructed two datasets, one for Fujifilm endoscopes and the other for Olympus endoscopes. We collected endoscope images capturing or inspecting gastric lesions, including early cancers, using the endoscopes of each manufacturer. The Fujifilm dataset consists of 1,063 SE images as the domain $X$ and 1,216 CE images, from which ECE images are generated as the domain $Y$. For the evaluation of the V-ECE image generation, we randomly selected 15 SE images as the input for testing. Similarly, the Olympus dataset consists of 1,064 SE images and 818 CE images. For testing, we randomly selected 14 SE images.

\subsection{Implementation Details}

We conducted two types of experiments to evaluate our tunable CycleGAN model: (i)~V-ECE image generation with only gradient enhancement (one-parameter model), and (ii)~V-ECE image generation with both gradient and color enhancements (two-parameter model). For the one-parameter model, the color gain $N_c$ was fixed to 1.2 and the gradient gain $N_g$ was randomly sampled from the discrete gain set of~$\left[1.0, 2.0, \cdots, 9.0\right]$. For the two-parameter model, the color gain $N_c$ was also randomly sampled from the discrete gain set of~$\left[1.0, 1.1, \cdots, 1.8\right]$. To construct the gain maps, we normalized each gain range from $[1.0,9.0]$ or $[1.0,1.8]$ to $[0,255]$ by applying a linear transformation, so that the generator and the discriminator do not neglect the small differences of the original values. For each experiment, we separately learned our tunable CycleGAN models for the Fujifilm and the Olympus datasets, considering that the internal image processing pipeline of the endoscope system depends on each manufacturer.

Regarding the training settings, we followed the default settings of the original CycleGAN implementation. The batch size was set to 1, and the number of epochs was set to 200. The learning rate was 0.0002 for the first 100 epochs and linearly decayed from 0.0002 to 0 during the remaining 100 epochs. The loss weights were set as $\lambda_\text{cyc} = 10$ and $\lambda_\text{identity} = 5$. The training was conducted using a single NVIDIA H100 GPU and took approximately two days.

\subsection{Results of One-Parameter Tunable Model}

We first evaluate the results of the one-parameter model. In addition to a visual evaluation, we perform a numerical evaluation using this model, because it is simpler than the two-parameter model, making the numerical results more interpretable.

Figure~\ref{fig:T-VEIC_visual_results} shows the translation results of our tunable CycleGAN model for test SE images in the Fujifilm (top) and the Olympus (bottom) datasets. We can confirm that the enhancement level is gradually increased according to the input gains at inference time, despite the ECE images being generated from a single model.

\begin{figure}[t!]
    \centering
    \includegraphics[width=1\linewidth]{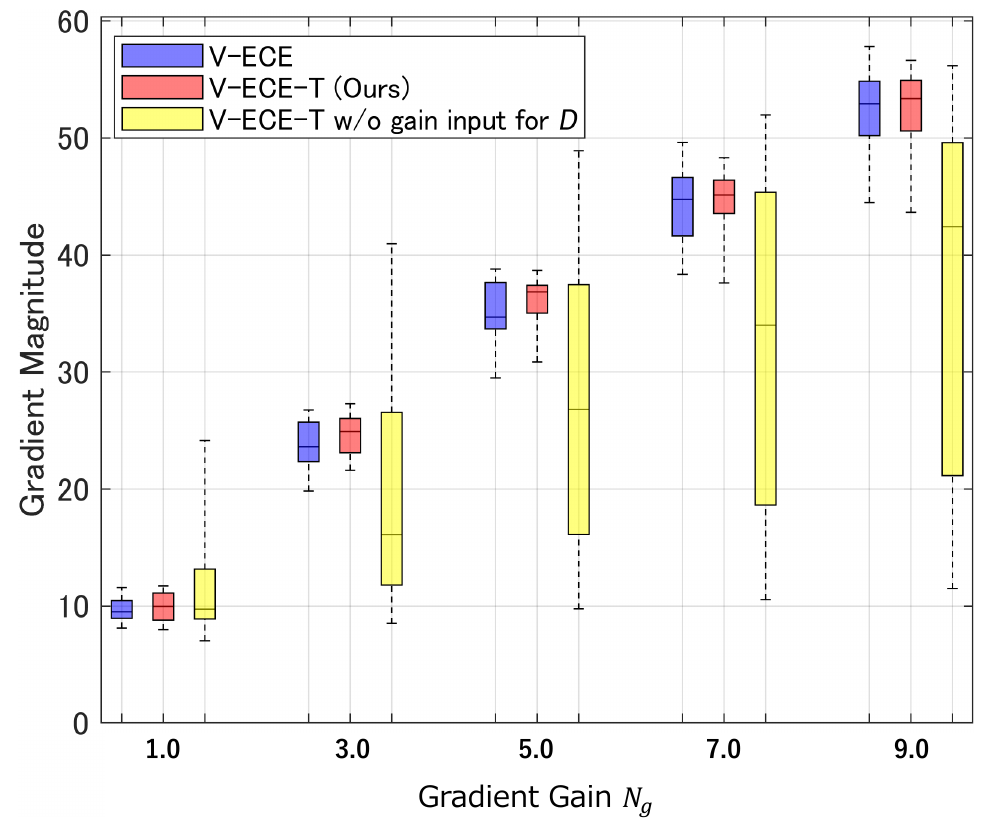}
    \caption{The numerical evaluation results of the V-ECE image generation. Our tunable model (V-ECE-T) generates consistent results with the independent models for each gain (V-ECE).}
    \label{fig:quantitative_results}
\end{figure}

We then perform a numerical evaluation. To assess the validity of our tunable CycleGAN model, we evaluate the consistency between our single tunable CycleGAN model and independent CycleGAN models trained for each gain (fixed for all training images). Figure~\ref{fig:quantitative_results} shows the results using the 15 test SE images of the Fujifilm dataset, where the vertical axis represents the boxplot of the gradient magnitude, which is calculated as  
\begin{equation}
    q_m(i,j) = \sqrt{(\partial_hu(i,j))^2+(\partial_vu(i,j))^2}.
    \label{eq:gradient_calc}
\end{equation}
Here, $u(i,j)$ is the pixel value of the generated V-ECE image at pixel $(i,j)$, and ($\partial_h$,$\partial_v$) are partial derivative operators in the horizontal and the vertical directions, respectively. In Fig.~\ref{fig:quantitative_results}, V-ECE represents the results of the independent CycleGAN models for each gain, presenting the reference results for the gradient magnitudes. V-ECE-T represents the results of our tunable CycleGAN model. We can confirm that our tunable CycleGAN can generate consistent results with the independent CycleGAN models, which validates the reasonableness of our results. As an ablation study, we also evaluate a tunable model, where the gain maps are inputted only for the generator, but not for the discriminator (V-ECE-T w/o gain input for $D$). This model generates a very diverse range of gradient magnitudes and significantly inconsistent results with the references (V-ECE), highlighting the importance of the gain map inputs to the discriminator.

\subsection{Results of Two-Parameter Tunable Model}

\begin{figure}[t!]
    \centering
    \begin{minipage}{1\linewidth}
        \centering
        \includegraphics[width=0.3\linewidth,
        trim=10mm 100mm 10mm 10mm,
        clip
        ]{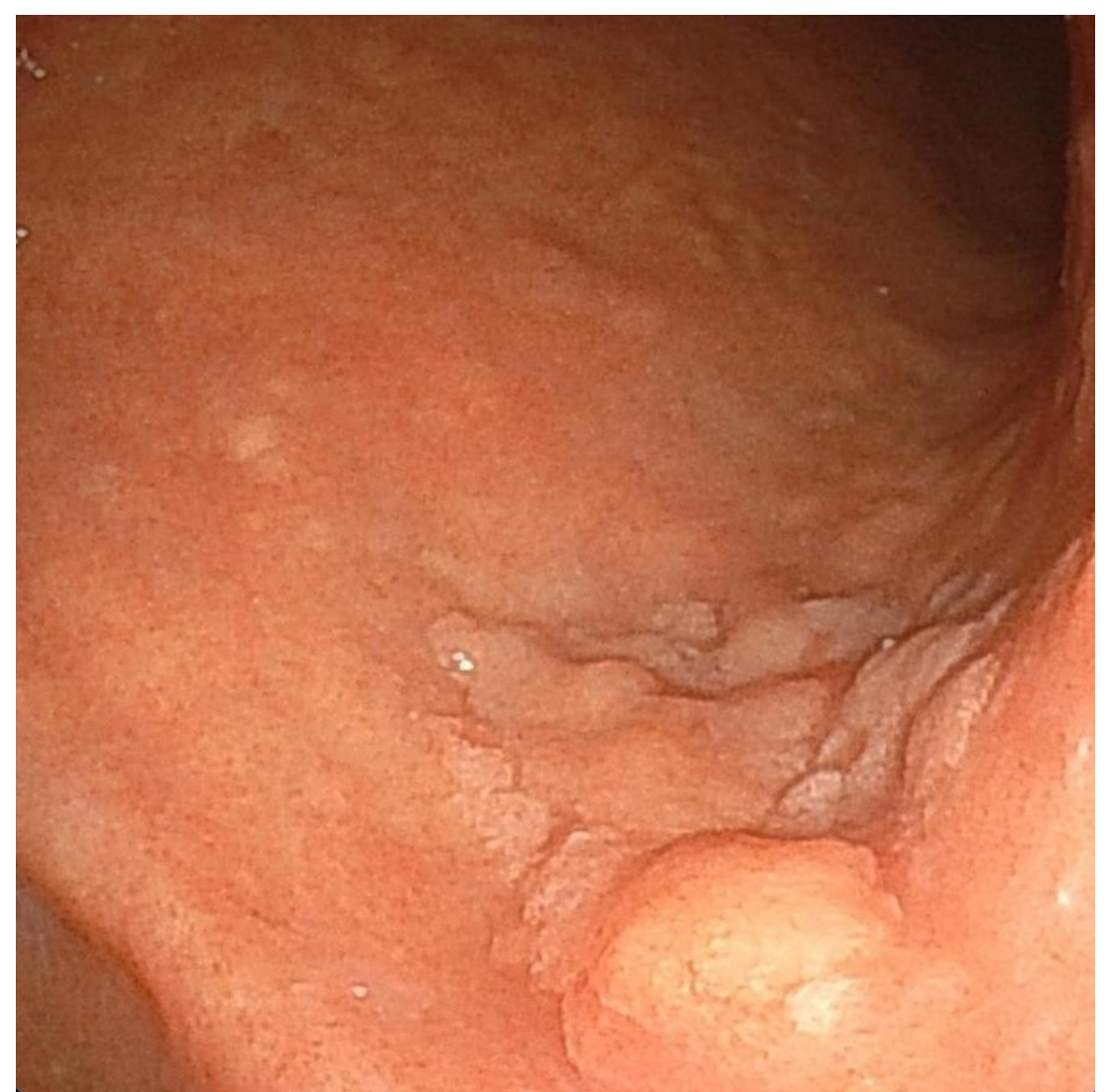}
        \subcaption{Input SE image}
    \end{minipage}
    \begin{minipage}{1\linewidth}
        \centering
        \includegraphics[width=1\linewidth]{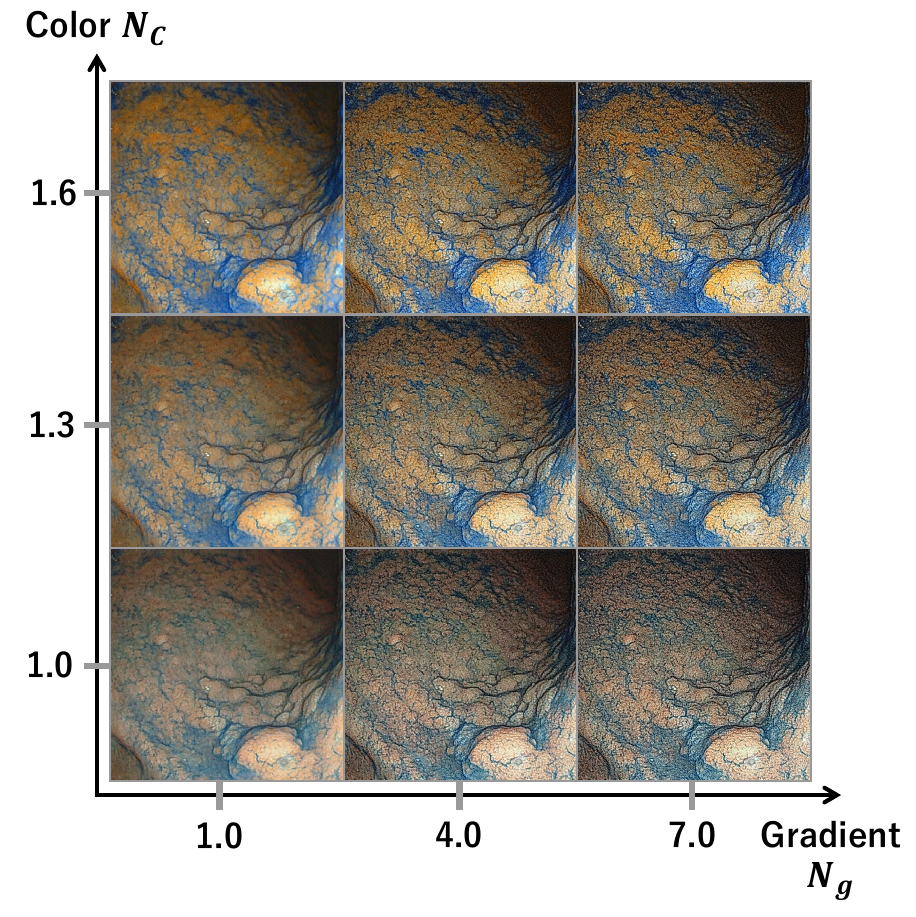}\\ \vspace{-2mm}
        \subcaption{Generated V-ECE images with each gain pair}
    \end{minipage} 
    \caption{The results of our tunable CycleGAN with the two-parameter model.}
    \label{fig:2_parameters_tunable}
\end{figure}

Figure~\ref{fig:2_parameters_tunable} shows the visual results of our tunable CycleGAN with the two-parameter model for one of the test SE images in the Fujifilm dataset. We can confirm that the strengths of the image gradients are gradually increased along the horizontal axis, while the levels of the color enhancements are gradually increased along the vertical axis. These results validate the effectiveness of our tunable CycleGAN, allowing us to jointly tune the gradient and color gains.

More results can be seen on our project page\footnote{\url{http://www.ok.sc.e.titech.ac.jp/res/VIC/}}.

\section{Conclusion}

In this paper, we have proposed V-ECE with a tunable CycleGAN model. Compared with the existing V-CE, our V-ECE adds the following values: (i) It generates enhanced images with improved visibility through gradient and color enhancements. (ii) Its enhancement level is tunable by a medical practitioner in the application phase. These factors improve the practicability of the virtual system for CE. Our tunable CycleGAN can be implemented in real time by replacing the model parameters of our current V-CE system~\cite{takasu2025assessment,suzuki2026generative} with those of the tunable CycleGAN. In future work, we plan to assess our V-ECE system from clinical aspects in real endoscopy environments.

\bibliographystyle{IEEEtran}
\bibliography{egbib}

\section*{Appendix}

This appendix explains the derivation of the fixed value $\alpha$ in Eq.~(\ref{eq:Cb_modify}). The conversion from YCbCr to RGB is described as
\begin{align}
    \begin{bmatrix}
        R\\G\\B
    \end{bmatrix}
    =T\begin{bmatrix}
        Y-16\\Cb-128\\Cr-128
    \end{bmatrix},
    \label{eq:ycbcr2rgb}
\end{align}
where $T$ is the conversion matrix~\cite{poynton1996technical} as below.

\begin{equation}
    T = \begin{bmatrix}
        1.164&0&1.596\\
        1.164&-0.391&-0.813\\
        1.164&2.018&0
    \end{bmatrix}.
\end{equation}
We let $[G, Cb, Cr]$ be the original pixel values before the color enhancement and $[G', Cb', Cr']$ be the pixel values after the enhancement. Then, the change of the G value, $\Delta G = G' -G$, is described as 
\begin{align}
    \Delta G &= T_{22}\Delta Cb + T_{23}\Delta Cr,
\end{align}
where $T_{nm}$ is the element of the matrix $T$ at $n$-th row and $m$-th column, and $\Delta Cb = Cb' -Cb$ and $\Delta Cr = Cr' -Cr$, respectively.
When we impose the condition that the G value is constant before and after the color enhancement, i.e., $\Delta G = 0$, we obtain the following equations.
\begin{align}
    0 &= T_{22}\Delta Cb + T_{23}\Delta Cr,\\
    \Delta Cb &= -T_{23}/T_{22}\Delta Cr.
    \label{eq:delta_cb}
\end{align}
Because we design $Cr' = N_c \cdot (Cr - 128) +128$ as in Eq.~(\ref{eq:cr'}), $\Delta Cr$ is described as
\begin{equation}
    \begin{aligned}
        \Delta Cr&=Cr'-Cr\\
        &=N_c(Cr-128)+128-Cr\\
        &=(N_c-1)(Cr-128).
    \end{aligned}
    \label{eq:delta_cr}
\end{equation}
According to Eqs.~(\ref{eq:delta_cb}) and (\ref{eq:delta_cr}), we finally obtain $Cb'$ as
\begin{align}
    \Delta Cb &= -T_{23}/T_{22}(N_c-1)(Cr-128), \\
    Cb'&= \alpha(N_c-1)(Cr-128)+Cb,
\end{align}
where $\alpha = -T_{23}/T_{22} \approx 2.08$ is a fixed constant.

\end{document}